\RequirePackage{fix-cm}
\documentclass[10pt,twocolumn]{ICCAS}

\usepackage{microtype}
\usepackage{xcolor}
\usepackage{graphicx}
\usepackage{subcaption}
\usepackage{cite}
\usepackage{caption}
\usepackage[a4paper, left=2cm, right=2cm, top=2.5cm, bottom=2.5cm]{geometry}
\usepackage{amssymb}
\usepackage{amsmath}
\usepackage{comment}

\graphicspath{{./Images/}}

\makeatletter
\def\bstctlcite#1{\@bsphack
  \@for\@citeb:=#1\do{%
    \edef\@citeb{\expandafter\@firstofone\@citeb}%
    \if@filesw
      \immediate\write\@auxout{\string\citation{\@citeb}}%
    \fi}%
  \@esphack}
\makeatother

\AtBeginDocument{%
  \setlength{\abovedisplayskip}{3pt}%
  \setlength{\belowdisplayskip}{3pt}%
  \setlength{\abovedisplayshortskip}{0pt}%
  \setlength{\belowdisplayshortskip}{3pt}%
}

\definecolor{revisedblue}{RGB}{0,82,204}

\makeatletter

\def\section{\@startsection {section}{1}{\z@}%
  {1.2ex plus 0.2ex minus .1ex}%
  {0.3ex}%
  {\center\fontsize{11pt}{11pt}\rm\bf\MakeUppercaseBlue}}

\def\subsection{\@startsection{subsection}{2}{\z@}%
  {0.7ex plus 0.1ex minus .1ex}%
  {0.2ex}%
  {\noindent\rm\bf\textcolor[RGB]{0,112,192}}}

\def\subsubsection{\@startsection{subsubsection}{3}{\z@}%
  {0.7ex plus 0.1ex minus .1ex}%
  {0.1ex}%
  {\rm}}

\makeatother

\usepackage{diagbox}

\begin{document}
\bstctlcite{BSTcontrol}

© 2026 IEEE. Personal use of this material is permitted.
Permission from IEEE must be obtained for all other uses,
including reprinting/republishing this material for advertising
or promotional purposes, collecting new collected works
for resale or redistribution to servers or lists, or reuse of
any copyrighted component of this work in other works.
This work has been submitted to the IEEE for possible
publication. Copyright may be transferred without notice,
after which this version may no longer be accessible.

\newpage

\title{Actuator-Aware Inverse Kinematics with Joint-Limit Admissibility for Torque-Controlled Redundant Robots}

\author{Mohammad Dastranj${}^{1*}$, Mahdi Hejrati${}^{1}$ and Jouni Mattila${}^{1}$ }

\affils{ ${}^{1}$Unit of Automation Technology and Mechanical Engineering, Faculty of Engineering and Natural Sciences \\
Tampere University, 33720 Tampere, Finland (mohammad.dastranj@tuni.fi) \\
{\small${}^{*}$ Corresponding author}}

\abstract{
    This paper proposes actuator-aware inverse kinematics for torque-controlled redundant robots under joint-limit constraints. In the considered architecture, the inverse-kinematic output is not merely a purely kinematic joint-velocity command; it is the required joint velocity supplied to a downstream torque-level controller. Therefore, a small commanded task residual may not necessarily improve realized motion. The proposed method formulates a convex quadratic programming problem whose decision variable is the joint-level required velocity. Control barrier function style bounds impose reference-level joint-limit admissibility, while the task equation is handled through a penalized slack variable. Redundancy is resolved using a controller-compatibility objective that accounts for previous-command consistency and actuator torque-capacity weighting. The method is independent of the particular torque-level controller and can serve as an intermediate IK layer between an endpoint trajectory and a redundant robot controller. Experiments on a virtual-decomposition-controlled seven-degree-of-freedom upper-limb exoskeleton compare the method with standard inverse-kinematic baselines and a constrained task-preserving quadratic programming baseline. The results indicate lower limit-pushing commands, bounded admissible required velocities, and improved realized task behavior in the tested trajectory, without modifying the downstream controller.
}

\keywords{
    inverse kinematics, torque-controlled redundant robots, quadratic programming, joint-limit constraints, controller-compatible reference generation.
}

\maketitle


\section{Introduction}

Redundant robots have more actuated joints than are required to realize a given task-space motion. This property gives them extra degrees of freedom that can be used to satisfy additional requirements while preserving the task whenever feasible. In such robots, the control objective is not typically limited to task performance, but may also involve physical interaction requirements, joint or state constraints, and actuator limitations \cite{proietti2016,hejrati2024}. Some upper-limb exoskeletons, such as the seven-degree-of-freedom (7-DoF) experimental platform in this research, are one example of such systems. In redundant robots, multiple joint-velocity commands may correspond to the same endpoint motion.

This redundancy increases dexterity \cite{martinez2023} and can be used to avoid certain restrictions, such as joint and actuator limits \cite{elias2024}. However, when the robot is controlled through a downstream torque-level controller, the selected inverse-kinematic (IK) solution is not merely a purely kinematic velocity command. It provides a joint-level desired motion from which the controller constructs the tracking terms used in the control law. For example, in virtual decomposition control (VDC), the desired motion is combined with pose tracking error to form the corresponding required velocities from which the required torques are computed \cite{zhu2010vdc,hejrati2022}. Therefore, the IK mapping should not only provide a task-preserving velocity solution, but should also account for how compatible the resulting reference is with the downstream controller.

Conventional redundant IK mappings are typically formulated at the velocity level by solving the inverse differential kinematics of the task Jacobian \cite{siciliano2009robotics,chiaverini2008,colome2015}. The pseudoinverse gives the minimum-norm least-squares joint velocity, while the general redundant solution adds a null-space velocity component for secondary objectives, such as joint-limit avoidance, without affecting the primary task whenever feasible \cite{siciliano2009robotics,chiaverini2008}. Near singular configurations, damped least-squares mappings regularize the inverse by relaxing task-space accuracy in favor of bounded joint-velocity commands \cite{chiaverini2008,colome2015}. Thus, these schemes resolve redundancy through kinematic command-level criteria: minimum norm selection, null-space projection, or singularity damping. Compatibility with downstream torque-level control and actuation constraints is not handled explicitly within the IK problem itself.

Recent redundant IK formulations often use constrained optimization, such as quadratic programming (QP) or related linear-programming formulations, to encode human interaction \cite{tassi2022}, joint motion limits \cite{faroni2020}, real-time inequality constraints \cite{zhang2024}, whole-body collision avoidance \cite{zhang2026}, manipulability maximization \cite{colan2023}, and similar objectives in a unified manner. Compared with standard null-space projection, these formulations provide a more direct way to impose constraints and handle competing objectives. In a related direction, control barrier functions (CBFs) provide a formal way to enforce safety-related forward-invariance conditions \cite{ames2017}, often through QP, and have been widely adopted in safety-critical control \cite{ames2019,basso2020}. In velocity-level IK, CBF-style inequalities can serve as limit-aware filters on commanded joint velocities. Their plant-level interpretation, however, depends on the control architecture. For velocity-controlled systems, the IK-generated joint velocity may be the control input, and such constraints can directly support joint-limit forward invariance. For torque-controlled systems, the IK output is only a reference to the downstream controller, while the physical input is actuator torque. Hence, without a closed-loop dynamic CBF proof, the constraints are interpreted as reference-level joint-limit awareness rather than plant-level safety guarantees.

In this paper, we formulate controller-compatible IK for torque-controlled redundant robots as an intermediate reference-generation map from task-space commands to joint-level required velocities for torque-level control. In this study, VDC is used as the downstream controller of the redundant upper-limb exoskeleton to compare different IK redundancy-resolution strategies under the same torque-level control framework. The proposed method formulates a QP whose decision variable is the joint-level reference velocity and whose CBF-style joint-limit constraints impose reference-level admissibility. The task equation is imposed as a soft constraint by introducing a slack variable, rather than using the differential task-kinematic residual directly as the main objective. This preserves the task at the reference level while preventing the task residual from overriding controller-compatibility considerations. The controller-compatibility objective includes joint-level terms relevant to the downstream torque-level controller, namely a reference-velocity step objective and an actuator torque-capacity objective based on known torque limits.

The proposed method is compared with pseudoinverse IK, damped least-squares IK, null-space joint-limit-avoidance IK, and a task-preserving QP IK. The task-preserving QP IK baseline uses the same reference-level admissibility constraints as the proposed method, isolating the effect of the controller-compatibility objective from constraint enforcement alone. The results show that the proposed formulation achieves a better balance between admissibility and task preservation: it avoids the limit-related degradation observed in the conventional IK baselines while preserving the realized task more effectively than the constrained task-preserving QP IK. This supports the central premise that, in a torque-controlled redundant robot, IK should not only reduce commanded kinematic error, but also generate joint-level references that remain compatible with the downstream controller.


\section{Controller-Compatible Redundant Inverse Kinematics}

For a redundant torque-controlled robot, let the required task velocity be $\boldsymbol{V}_r\in{\mathbb{R}}^{m}$ and the required joint velocity be $\boldsymbol{\dot q}_r\in{\mathbb{R}}^{n}$, with $n>m$. When the task Jacobian $\boldsymbol{J}(\boldsymbol{q})\in{\mathbb{R}}^{m\times n}$ has full row rank, the differential kinematics
\begin{equation}
    \boldsymbol{V}_r
    =
    \boldsymbol{J}(\boldsymbol{q})\boldsymbol{\dot q}_r
    \label{eq:differentialKinematics}
\end{equation}
has $n-m$ degrees of redundancy. Therefore, multiple required joint velocities may represent the same task velocity when no additional restrictions are active. Eq.~(\ref{eq:differentialKinematics}) defines the nominal task-velocity relation used by all IK formulations. In a torque-controlled robot, however, $\boldsymbol{\dot q}_r$ is a required-velocity reference supplied to the downstream torque-level controller; therefore, minimizing the commanded residual $\left\|\boldsymbol{V}_r-\boldsymbol{J}(\boldsymbol{q})\boldsymbol{\dot q}_r\right\|$ alone does not ensure good realized motion when the resulting reference is difficult to track. This motivates resolving redundancy using not only task preservation and joint-limit admissibility, but also controller compatibility.

\subsection{Joint-Limit-Admissible Required-Velocity Set}

Joint $i$ position bounded by lower and upper limits $\underline{q}_{\,i}$ and $\bar{q}_i$, respectively, is subject to the inequality constraint
\begin{equation}
    \underline{q}_{\,i} \le q_i \le \bar{q}_i.
\end{equation}
Using the standard CBF condition \cite{ames2017} for this constraint and replacing the actual joint velocities by $\dot{q}_{r,i}$, the corresponding CBF-style admissible required-velocity bounds become
\begin{equation}
    \dot{q}_{r,i}
    \ge
    -\alpha_i\Bigl(q_i-\underline{q}_{\,i}\Bigr),
    \label{eq:lowerCBFStyleBound}
\end{equation}
\begin{equation}
    \dot{q}_{r,i}
    \le
    \;\,\alpha_i\Bigl(\bar{q}_i-q_i\Bigr),
    \label{eq:upperCBFStyleBound}
\end{equation}
where \(\alpha_i>0\) determines the allowed approach rate toward the joint limits. These bounds are intersected with the prescribed required velocity limits
\(\underline{\dot{q}}_{\,r,i}\) and \(\bar{\dot{q}}_{r,i}\), giving
\begin{equation}
    \dot{q}_{r,\mathrm{lb},i}(q_i)
    =
    \max\Bigl\{
    \underline{\dot{q}}_{\,r,i},
    -\alpha_i\Bigl(q_i-\underline{q}_{\,i}\Bigr)
    \Bigr\},
    \label{eq:qdotLowerBound}
\end{equation}
\begin{equation}
    \dot{q}_{r,\mathrm{ub},i}(q_i)
    =
    \min\Bigl\{
    \bar{\dot{q}}_{\,r,i},\;\,
    \alpha_i\Bigl(\bar{q}_i-q_i\Bigr)
    \Bigr\}
    \label{eq:qdotUpperBound}
\end{equation}
for the lower- and upper-bounds on the admissible required joint velocities. The admissible required-velocity set is then
\begin{equation}
    {\mathcal{D}}_r(\boldsymbol{q})
    =
    \left\{
    \dot{\boldsymbol{q}}_r\in{\mathbb{R}}^{n}
    \;\middle|\;
    \dot{\boldsymbol{q}}_{r,\mathrm{lb}}(\boldsymbol{q})
    \le
    \dot{\boldsymbol{q}}_r
    \le
    \dot{\boldsymbol{q}}_{r,\mathrm{ub}}(\boldsymbol{q})
    \right\},
    \label{eq:admissibleSet}
\end{equation}
where the inequalities are understood elementwise. The constraints are called CBF-style because they have the same first-order joint-limit form as CBF velocity inequalities, but they are imposed on the required joint velocities generated by the IK layer rather than directly on the plant input.

\subsection{Controller-Compatibility Cost}
Within the admissible set, the proposed IK solver resolves redundancy using a controller-compatibility cost. The proposed weighting matrix is
\begin{equation}
    \boldsymbol{W}_{\!\!c}
    =
    \boldsymbol{W}_{\!\!0}
    +
    \boldsymbol{W}_{\!\!\!\Delta}
    +
    \boldsymbol{W}_{\!\!\tau}.
    \label{eq:compatibilityWeighting}
\end{equation}
Here, $\boldsymbol{W}_{\!\!0}=w_0\boldsymbol{I}_n$ with $w_0>0$, is a small isotropic baseline weighting. It keeps the quadratic cost well-conditioned and provides neutral minimum-norm behavior when the compatibility terms are weak.

The first active term, $\boldsymbol{W}_{\!\!\!\Delta}$, discourages abrupt changes in the required velocity supplied to the downstream controller. Let $\boldsymbol{\dot q}_{r,k-1}$ denote the required joint velocity vector from the previous step, and let $\Delta_{\mathrm{ref},i}>0$ be a reference step scale for joint $i$. The required-velocity step weighting is
\begin{equation}
    \boldsymbol{W}_{\!\!\!\Delta}
    =
    \mathrm{diag}
    \left(
    w_{{\scriptscriptstyle \Delta},1},\ldots,w_{{\scriptscriptstyle \Delta},n}
    \right),
\end{equation}
with
\begin{equation}
    w_{{\scriptscriptstyle \Delta},i}
    =
    \frac{\rho_{{\scriptscriptstyle \Delta}}}
    {\Delta_{\mathrm{ref},i}^{2}+\varepsilon_{\scriptscriptstyle W}},
    \label{eq:stepWeight}
\end{equation}
where $\rho_{{\scriptscriptstyle \Delta}}>0$ is a weighting gain and $\varepsilon_{{\scriptscriptstyle W}}>0$ is to avoid division by zero. $\boldsymbol{W}_{\!\!\!\Delta}$ also defines the previous-command consistency vector
\begin{equation}
    \boldsymbol{h}
    =
    \boldsymbol{W}_{\!\!\!\Delta}\,
    \boldsymbol{\dot q}_{r,k-1}.
    \label{eq:previousCommandConsistency}
\end{equation}
Thus, proximity to the previous required velocity is introduced as a soft optimization preference, not as a hard rate limiter.

The second active term, $\boldsymbol{W}_{\!\!\tau}$, discourages excessive reliance on low-capacity actuators when resolving the redundancy. Let \(\bar{\tau}_i>0\) denote the usable actuator torque capacity of joint \(i\). The normalized capacity is
\begin{equation}
    c_i
    =
    \frac{\bar{\tau}_i}
    {\max_j \bar{\tau}_j}.
    \label{eq:normalizedCapacity}
\end{equation}
The denominator uses the maximum torque capacity among all joints, where \(j\) is a dummy joint index. Thus, \(c_i\in(0,1]\) expresses the torque capacity of joint \(i\) relative to the strongest joint. If separate positive and negative torque limits are available, \(\bar{\tau}_i\) may be chosen as the smaller absolute limit to obtain a conservative symmetric capacity. The actuator-capacity weighting is
\begin{equation}
    \boldsymbol{W}_{\!\!\tau}
    =
    \mathrm{diag}
    \left(
    w_{\tau,1},\ldots,w_{\tau,n}
    \right),
\end{equation}
where
\begin{equation}
    w_{\tau,i}
    =
    \rho_{\tau}\eta_i,
    \label{eq:torqueWeight}
\end{equation}
and
\begin{equation}
\eta_i
=
\min\left\{
\bar{\eta},
\max\left\{
0,
\frac{1}{c_i^2+\varepsilon_{\scriptscriptstyle W}}-1
\right\}
\right\}.
    \label{eq:torquePressure}
\end{equation}
Here, \(\eta_i\) is a normalized capacity penalty factor for joint \(i\). Since \(c_i=1\) for the strongest joint, the term \((c_i^2+\varepsilon_{\scriptscriptstyle W})^{-1}-1\) is approximately zero for the strongest joint and increases as the available torque capacity decreases. Therefore, lower-capacity joints receive larger weights in the IK cost and become less favorable directions for generating the required velocity. The upper bound \(\bar{\eta}\) prevents excessive weighting differences between joints, and \(\rho_{\tau}>0\) scales the overall influence of the actuator-capacity term. This weighting depends only on known actuator limits and is not computed from measured torque.

\subsection{Controller-Compatible IK-QP Formulation}

To maintain feasibility when the task command is incompatible with the admissible required-velocity set, the task relation is selected to be a soft constraint by introducing the slack variable $\boldsymbol{s}\in{\mathbb{R}}^{m}$ instead of being an optimization objective:
\begin{equation}
    \boldsymbol{J}(\boldsymbol{q})\boldsymbol{\dot q}_r
    +
    \boldsymbol{s}
    =
    \boldsymbol{V}_r .
    \label{eq:slackenedTaskConsistency}
\end{equation}
Using Eq. (\ref{eq:slackenedTaskConsistency}) and the admissible set in Eq. (\ref{eq:admissibleSet}), the proposed controller-compatible IK-QP is formulated as
\begin{equation}
\begin{aligned}
\min_{\boldsymbol{\dot q}_r,\boldsymbol{s}}
\quad
&
\frac{1}{2}
\boldsymbol{\dot q}_r^{T}
\boldsymbol{W}_c
\boldsymbol{\dot q}_r
-
\gamma
\boldsymbol{h}^{T}
\boldsymbol{\dot q}_r
+
\rho_s
\boldsymbol{s}^{T}
\boldsymbol{Q}_v
\boldsymbol{s}
\\
\mathrm{s.t.}
\quad
&
\boldsymbol{J}(\boldsymbol{q})\boldsymbol{\dot q}_r
+
\boldsymbol{s}
=
\boldsymbol{V}_r,
\\
&
\boldsymbol{\dot q}_r
\in
{\mathcal{D}}_r(\boldsymbol{q}).
\end{aligned}
\label{eq:proposedQP}
\end{equation}
In Eq. (\ref{eq:proposedQP}), $\boldsymbol{W}_c\succ0$ is the controller-compatibility weighting matrix, $\boldsymbol{h}$ is the previous-command consistency vector from Eq. (\ref{eq:previousCommandConsistency}), and $\gamma\in[0,1]$ sets the strength of this consistency term. The matrix $\boldsymbol{Q}_v\succ0$ is a chosen task-space weighting matrix for the slack penalty, and $\rho_s>0$ penalizes deviation from the nominal task-velocity relation in Eq. (\ref{eq:differentialKinematics}). The optimization is a convex QP because the objective is quadratic, the equality and bound constraints are affine, $\boldsymbol{W}_c\succ0$ by construction, and $\boldsymbol{Q}_v\succ0$ is chosen positive definite.

The objective in Eq. (\ref{eq:proposedQP}) balances task preservation and controller compatibility. The first two terms select a required joint velocity that is inexpensive under the controller-compatibility weighting and softly consistent with the previous required velocity. The slack term is used only to absorb task infeasibility caused by the admissibility constraints and is penalized so that the task is preserved as much as possible.

The constraint $\boldsymbol{\dot q}_r\in{\mathcal{D}}_r(\boldsymbol{q})$ enforces admissibility and boundedness of the required joint velocity generated by the proposed IK layer. The downstream controller is not modified by the IK formulation; only the IK map that generates $\boldsymbol{\dot q}_r$ is changed. Therefore, the generated signal remains the same type of required joint velocity supplied to the selected torque-level controller.


\section{Experiments and Results}
\begin{figure*}[tb]
    \centering
    \includegraphics[
        width=\linewidth,
        trim=3.4cm 5.2cm 3.5cm 5.9cm,
        clip
    ]{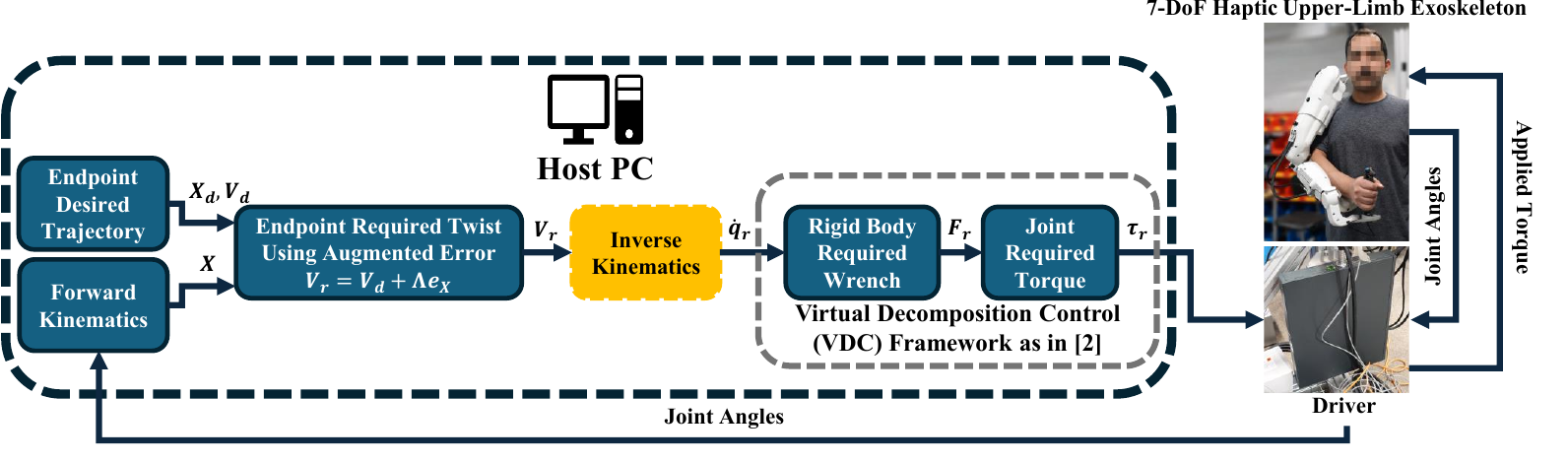}
    \caption{Experimental architecture for testing the redundant inverse kinematics. The exoskeleton communicates with the driver, which exchanges data with the host PC via EtherCAT. The VDC framework is kept as the torque-level controller. The inverse-kinematic layer is an independent intermediate block whose structure can affect downstream-controller behavior in a redundant robot.}
    \label{fig:Experiment}
\end{figure*}

In this paper, the downstream torque-level controller is kept the same as the VDC controller used in \cite{hejrati2024} for the 7-DoF upper-limb exoskeleton considered here. For this endpoint pose-tracking task, the required endpoint twist satisfies $\boldsymbol{V}_r\in{\mathbb{R}}^{6}$, whereas the exoskeleton has seven actuated joints, and hence $\boldsymbol{\dot q}_r\in{\mathbb{R}}^{7}$. The required endpoint twist construction, torque-level control law, subsystem decomposition, adaptation law, and virtual-stability argument therefore follow \cite{hejrati2024}. The present paper modifies only the redundant IK layer that maps the endpoint required twist $\boldsymbol{V}_r$ to the joint required velocity $\boldsymbol{\dot q}_r$; the VDC controller itself is not redesigned, and the purpose is to generate a more controller-compatible $\boldsymbol{\dot q}_r$ for the same downstream controller. The proposed IK method can also fit any other torque-level controller as the intermediate layer for mapping the endpoint trajectory to the joints. The experimental setup and the position of the IK layer in the VDC framework are shown in Fig.~\ref{fig:Experiment}. The host PC uses MatLab/Simulink R2024b and Virtuose interface, provided by Haption for the commercial ABLE upper-limb exoskeleton used in this research, to transfer the required data.

The experiments evaluate whether controller-compatible IK improves the motion realized by a torque-controlled redundant exoskeleton. The proposed method is compared with four baselines: minimum-norm Jacobian pseudoinverse (PINV), damped least-squares IK (DLS), null-space joint-limit-avoidance IK (NS-JLA), and a task-preserving QP baseline with CBF-style constraints, denoted by TP-QP. The closed-form baselines use standard differential IK and null-space projection forms \cite{chiaverini2008}. The task Jacobian is assumed to be full row rank in the considered configurations, so $\boldsymbol{J}\boldsymbol{J}^{T}$ is nonsingular for PINV.

For NS-JLA, the null-space joint-centering velocity command is
\begin{equation}
    \boldsymbol{\dot q}_{\mathrm{null}}
    =
    k_{\mathrm{null}}
    \boldsymbol{D}_{q}^{-1}
    \left(
    \boldsymbol{q}_{\mathrm{mid}}
    -
    \boldsymbol{q}
    \right),
    \label{eq:null_velocity_command}
\end{equation}
where
\begin{equation}
    \boldsymbol{q}_{\mathrm{mid}}
    =
    \frac{1}{2}
    \left(
    \boldsymbol{\bar{q}}
    +
    \boldsymbol{\underline{q}}
    \right),
    \qquad
    \boldsymbol{D}_{q}
    =
    \frac{1}{2}
    \operatorname{diag}
    \left(
    \boldsymbol{\bar{q}}
    -
    \boldsymbol{\underline{q}}
    \right).
    \label{eq:null_centering_terms}
\end{equation}
The TP-QP baseline is formulated as
\begin{equation}
\begin{aligned}
\min_{\boldsymbol{\dot q}_r} \quad &
\left(
\boldsymbol{J}\boldsymbol{\dot q}_r
-
\boldsymbol{V}_{\!\!r}
\right)^{T}
\boldsymbol{Q}_s
\left(
\boldsymbol{J}\boldsymbol{\dot q}_r
-
\boldsymbol{V}_{\!\!r}
\right)
\\
\mathrm{s.t.} \quad &
\boldsymbol{\dot q}_{r,\mathrm{lb}}(\boldsymbol{q})
\le
\boldsymbol{\dot q}_r
\le
\boldsymbol{\dot q}_{r,\mathrm{ub}}(\boldsymbol{q}) .
\end{aligned}
\label{eq:tp_qp}
\end{equation}
TP-QP uses the same required-velocity admissibility constraints as the proposed method but keeps the conventional task-residual objective, isolating the proposed controller-compatibility objective from admissibility enforcement alone.

All methods are tested using the same desired endpoint trajectory with similar initial configurations, the same VDC controller, and the same definition of the required task twist $\boldsymbol{V}_{\!\!r}$ as in \cite{hejrati2024}. Therefore, the comparison is made at the IK layer while the downstream torque-level controller remains unchanged. Because each IK output is supplied to the VDC controller as the required joint velocity, the evaluation includes both command-level kinematic consistency and closed-loop behavior: realized task tracking, realized task-velocity residuals, required-velocity admissibility, and near-limit commands. The experiment is not a plant-level joint-limit safety proof; it evaluates whether the IK-generated required velocity is more compatible with the unchanged VDC-controlled exoskeleton.

The numerical settings are summarized in Table~\ref{tab:exp_settings}. The mechanical joint limits and maximum allowable input torques are fixed platform-specific model values, used identically for all methods but omitted because of non-disclosure restrictions; all admissibility and limit-pushing metrics are computed with respect to these same internal limits. For fairness, TP-QP uses the same task-space residual metric as the slack penalty of the proposed method. The scalar slack weight $\rho_s = 10^4$ controls the proposed tradeoff between task preservation and controller-compatibility terms, whereas TP-QP contains only the task-residual objective, so any positive scalar scaling of $Q_s$ leaves its optimizer unchanged.

\begin{table}[b]
\centering
\caption{Numerical settings used for the IK comparison. Platform-specific joint and torque limits are fixed internally and omitted because of non-disclosure restrictions. Here, $a^{\times n}$ denotes $n$ consecutive entries equal to $a$.}
\label{tab:exp_settings}
\scriptsize
\setlength{\tabcolsep}{2pt}
\renewcommand{\arraystretch}{0.90}
\begin{tabular}{lp{0.825\columnwidth}}
\hline
Parameter & Value \\
\hline
$\Delta t$ & $10^{-3}\,\mathrm{s}$ \\
$\boldsymbol{Q}_v,\;\boldsymbol{Q}_s$ & $\boldsymbol{I}_6$ \\
$\boldsymbol{\dot q}_{r,\mathrm{lb/ub}}$ &
$\mp[2^{\times 4},1^{\times 3}]^T\,\mathrm{rad/s}$ \\
$\boldsymbol{\alpha}_{q}$ &
$[1^{\times 4},0.5^{\times 3}]^T$ \\
$\boldsymbol{d}$ &
$d_j=\pi/18\,\mathrm{rad}$, $j=1,\ldots,7$ \\
DLS & $\lambda=10^{-3}$ \\
NS-JLA &
$k_{\mathrm{null}}=0.8$; $\dot q_{\mathrm{null},\max}=0.8\,\mathrm{rad/s}$; $d_{\mathrm{null}}=\pi/9\,\mathrm{rad}$ \\
Proposed &
$\rho_{\Delta}=0.0035$; $\Delta_{\mathrm{ref},i}=0.06\,\mathrm{rad/s}$; $\gamma=0.15$; $\rho_{\tau}=0.30$; $\bar{\eta}=10$; $\rho_s=10^{4}$; $|s_i|\le 10^{3}$ \\
QP solver &
reg. $10^{-9}$; tol. $(10^{-8},10^{-8},10^{-10})$; max. iter. $400$ \\
\hline
\end{tabular}
\end{table}

Let the experimental samples be indexed by $k=1,\ldots,N$ at sampling instants $t_k$. The logged joint position, actual joint velocity, required joint velocity, and joint-input torque are $\boldsymbol{q}_k$, $\boldsymbol{\dot q}_k$, $\boldsymbol{\dot q}_{r,k}$, and $\boldsymbol{\tau}_k$, respectively, with $\boldsymbol{J}_k=\boldsymbol{J}(\boldsymbol{q}_k)$ and required task twist $\boldsymbol{V}_{r,k}$. RMS denotes root mean square.

The task-space tracking error $\boldsymbol{e}_{X,k}\in{\mathbb{R}}^{6}$ is decomposed into position and orientation components as
\begin{equation}
    \boldsymbol{e}_{p,k}
    =
    \left(\boldsymbol{e}_{X,k}\right)_{1:3},
    \qquad
    \boldsymbol{e}_{o,k}
    =
    \left(\boldsymbol{e}_{X,k}\right)_{4:6}.
\end{equation}
The reported tracking metrics, $E_{p,\mathrm{RMS}}$, $E_{o,\mathrm{RMS}}$, $E_{p,\max}$, and $E_{o,\max}$, are computed from $\|\boldsymbol{e}_{p,k}\|_2$ and $\|\boldsymbol{e}_{o,k}\|_2$ and are reported separately because of their different units.

Command-level IK consistency is evaluated using
\begin{equation}
    \boldsymbol{r}_{c,k}
    =
    \boldsymbol{J}_k
    \boldsymbol{\dot q}_{r,k}
    -
    \boldsymbol{V}_{r,k}.
    \label{eq:commanded_residual}
\end{equation}
This residual measures how closely the required joint velocity satisfies the instantaneous kinematic task equation. However, since $\boldsymbol{\dot q}_{r,k}$ is only a reference to the downstream torque-level VDC controller, a small commanded residual does not guarantee that the same task motion is realized near joint limits. The realized task-velocity residual is therefore
\begin{equation}
    \boldsymbol{r}_{a,k}
    =
    \boldsymbol{J}_k
    \boldsymbol{\dot q}_{k}
    -
    \boldsymbol{V}_{r,k}.
    \label{eq:actual_residual}
\end{equation}
The RMS linear and angular components of $\boldsymbol{r}_{a,k}$ are denoted by $r_{a,p,\mathrm{RMS}}$ and $r_{a,o,\mathrm{RMS}}$, respectively. Unlike the commanded residual, this metric evaluates the task velocity actually realized by the controlled exoskeleton.

Required-velocity admissibility is evaluated through the maximum bound violation. With the elementwise positive part denoted by $(\cdot)_+$,
\begin{equation}
    \epsilon_{\dot q,k}
    =
    \left\|
    \left(
    \boldsymbol{\dot q}_{r,k}
    -
    \boldsymbol{\dot q}_{r,\mathrm{ub}}
    \right)_+
    +
    \left(
    \boldsymbol{\dot q}_{r,\mathrm{lb}}
    -
    \boldsymbol{\dot q}_{r,k}
    \right)_+
    \right\|_{\infty}.
    \label{eq:qdot_bound_excess}
\end{equation}
The reported value is
\begin{equation}
    \epsilon_{\dot q,\max}
    =
    \max_k
    \epsilon_{\dot q,k}.
    \label{eq:max_qdot_bound_excess}
\end{equation}
It is zero only when the generated required joint velocity remains inside the admissible bounds at all samples.

\begin{figure}[tb]
\centering
\includegraphics[width=0.95\linewidth]{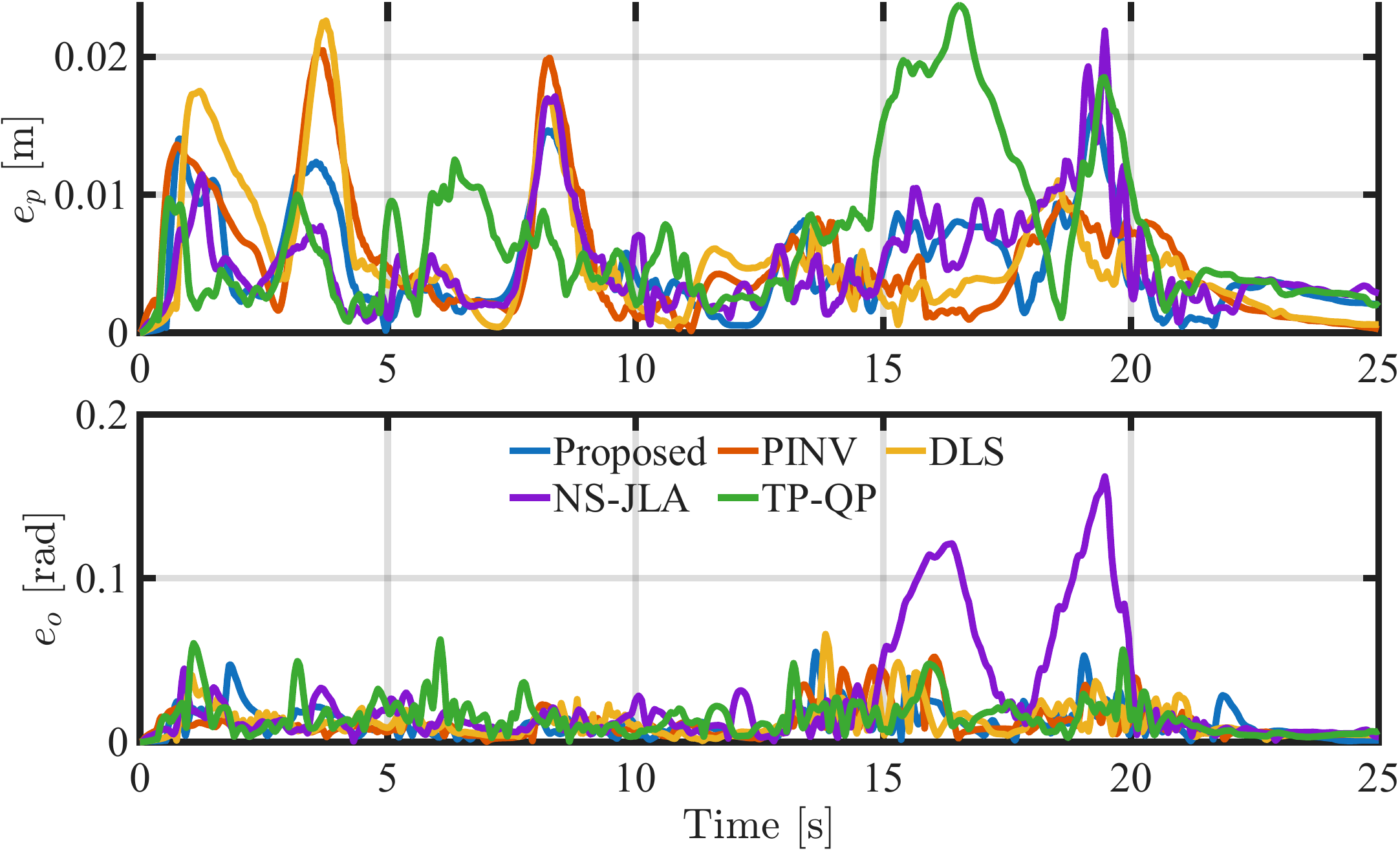}
\caption{Realized endpoint tracking errors obtained with the proposed method and the IK baselines, shown separately for position and orientation components.}
\label{fig:exp_tracking_error}
\end{figure}

\begin{figure}[tb]
\centering
\includegraphics[width=0.95\linewidth]{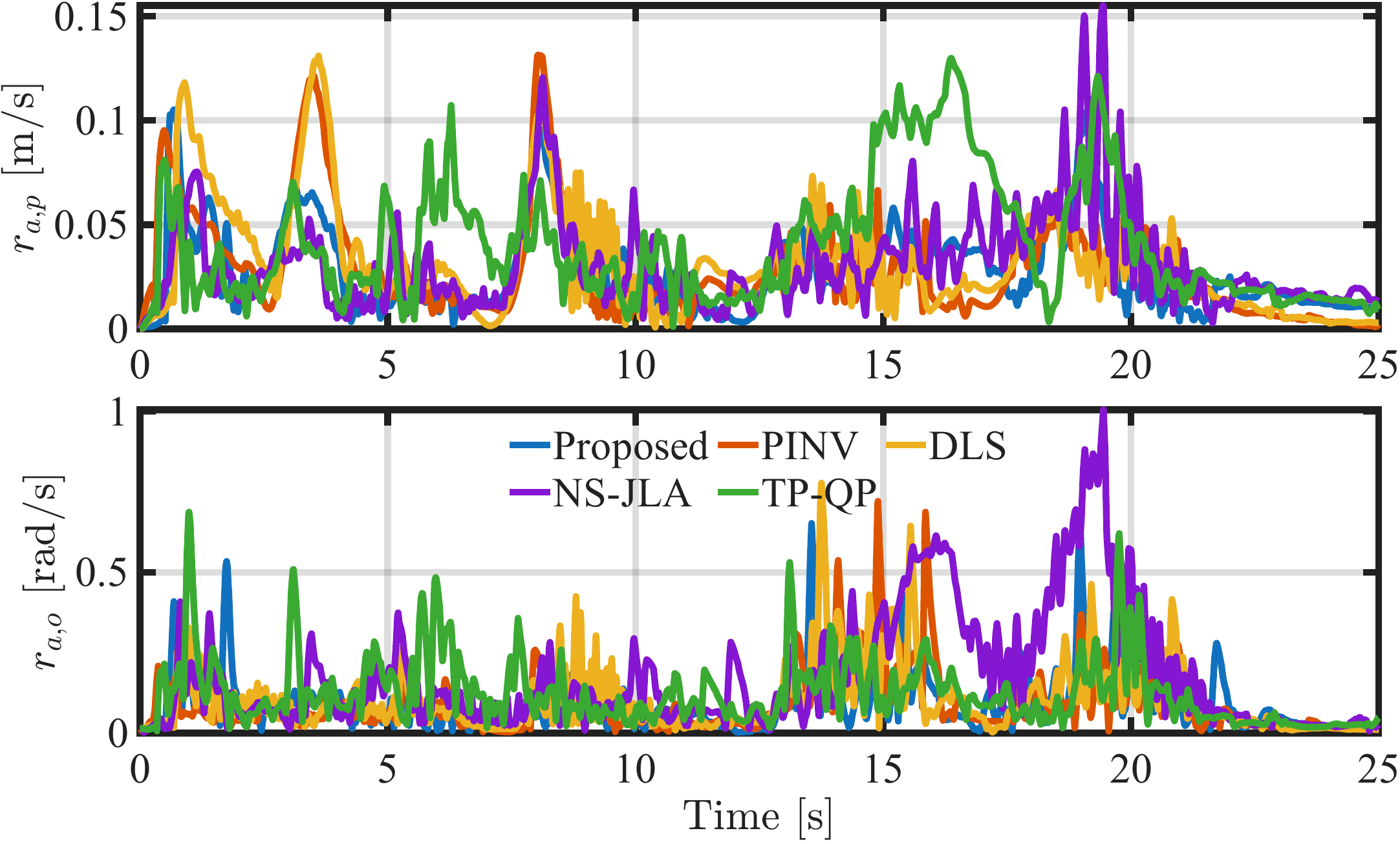}
\caption{Realized task-velocity residuals for all methods, computed from the actual joint velocity $\dot{q}$ rather than the commanded required velocity $\dot{q}_r$.}
\label{fig:exp_actual_residual}
\end{figure}

\begin{figure}[tb]
\centering
\includegraphics[width=0.95\linewidth]{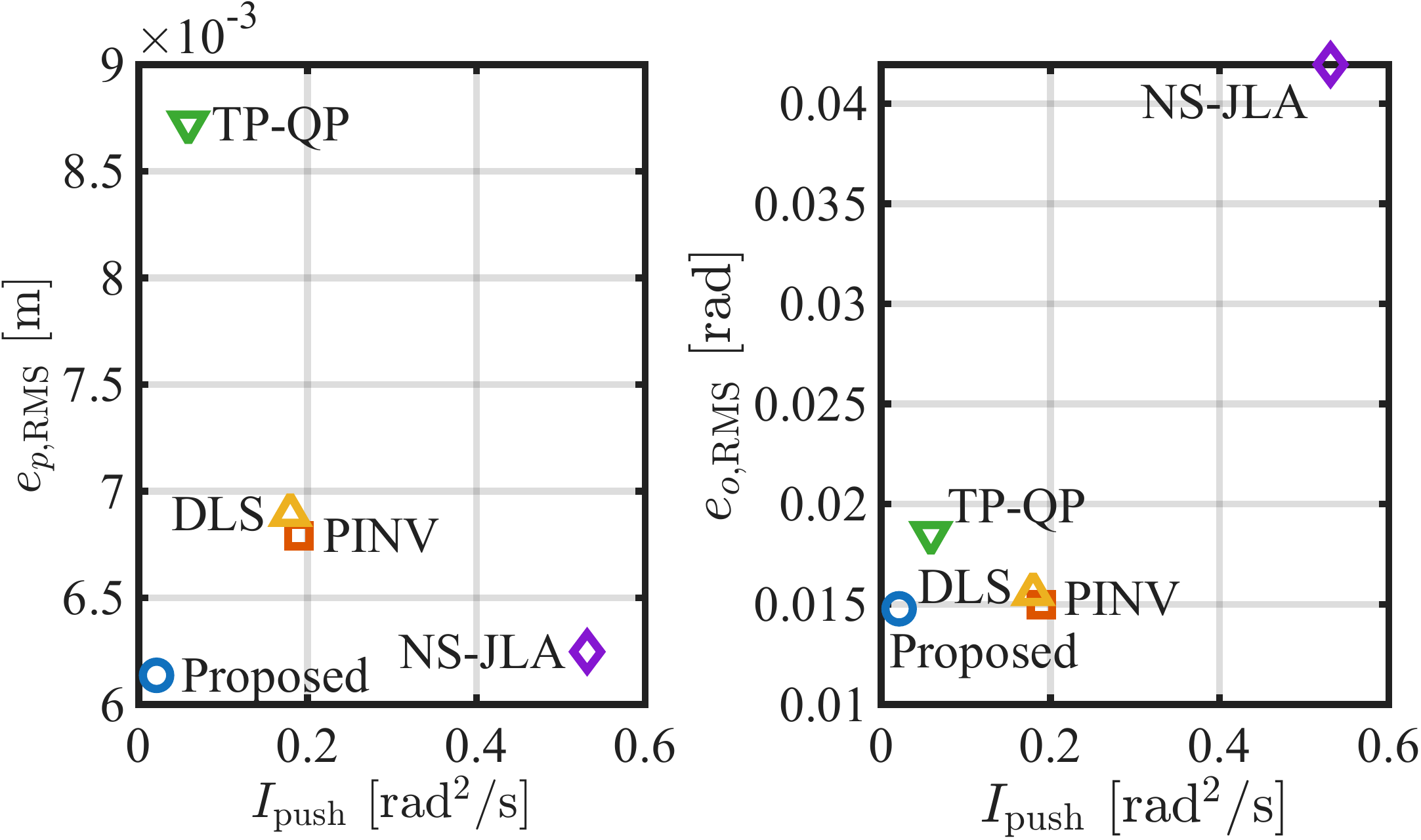}
\caption{Limit-pushing and realized tracking tradeoff for the proposed method and IK baselines. The proposed method remains in the low-pushing and low-error region.}
\label{fig:exp_pareto_push_tracking}
\end{figure}

For joint $j$, the lower and upper joint-limit clearances are
\begin{equation}
    m^{\underline q}_{j,k}
    =
    q_{j,k}
    -
    \underline q_{j},
    \qquad
    m^{\bar q}_{j,k}
    =
    \bar q_{j}
    -
    q_{j,k}.
\end{equation}
The limit-pushing command measures how much of the required velocity points toward a nearby mechanical limit:
\begin{equation}
p_{j,k}
=
\begin{cases}
\max(\dot q_{r,j,k},0),
&
m^{\bar q}_{j,k}\le d_j,
\\[1mm]
\max(-\dot q_{r,j,k},0),
&
m^{\underline q}_{j,k}\le d_j,
\\[1mm]
0,
&
\text{otherwise}.
\end{cases}
\label{eq:limit_pushing_component}
\end{equation}
where $d_j>0$ is the clearance threshold defining the near-limit region for joint $j$. With $P_k=\|\boldsymbol{p}_k\|_2$, the limit-pushing index is
\begin{equation}
    I_{\mathrm{push}}
    \approx
    \sum_{k=1}^{N-1}
    \frac{\Delta t_k}{2}
    \left(
    P_k^2
    +
    P_{k+1}^2
    \right),
    \label{eq:limit_pushing_index}
\end{equation}
where $\Delta t_k=t_{k+1}-t_k$. This index has units of $\mathrm{rad}^2/\mathrm{s}$ and quantifies commands requesting motion toward a nearby limit. Since a low $I_{\mathrm{push}}$ is desirable only without degrading realized task tracking, it is interpreted together with $E_{p,\mathrm{RMS}}$, $E_{o,\mathrm{RMS}}$, and the realized task-velocity residuals. The experimental comparison is summarized in Table~\ref{tab:main_exp_comparison} and Figs.~\ref{fig:exp_tracking_error}--\ref{fig:exp_pareto_push_tracking}, which report commanded IK consistency, realized task tracking, realized task-velocity residuals, limit-pushing behavior, required-velocity admissibility, and RMS torque norm.

\begin{table}[b]
\centering
\caption{Experimental comparison of the proposed method against the IK baselines. Lower values are better, and the best value in each row is shown in bold.}
\label{tab:main_exp_comparison}
\footnotesize
\setlength{\tabcolsep}{3pt}
\renewcommand{\arraystretch}{0.95}
\begin{tabular}{@{}lccccc@{}}
\hline
Metric & PINV & DLS & NS-JLA & TP-QP & Proposed \\
\hline
$r_{c,p,\mathrm{RMS}}$ [m/s]
& {\boldmath\textbf{$\approx 0$}} & 2.29e-6 & 3.46e-6 & 0.03385 & 0.01241 \\
$r_{c,o,\mathrm{RMS}}$ [rad/s]
& {\boldmath\textbf{$\approx 0$}} & 3.80e-7 & 9.29e-6 & 0.01196 & 0.00157 \\
$E_{p,\mathrm{RMS}}$ [m]
& 0.00679 & 0.00689 & 0.00625 & 0.00872 & \textbf{0.00614} \\
$E_{o,\mathrm{RMS}}$ [rad]
& 0.01499 & 0.01557 & 0.04194 & 0.01850 & \textbf{0.01477} \\
$E_{p,\max}$ [m]
& 0.02056 & 0.02261 & 0.02189 & 0.02373 & \textbf{0.01592} \\
$E_{o,\max}$ [rad]
& \textbf{0.05202} & 0.06594 & 0.16219 & 0.06258 & 0.05502 \\
$r_{a,p,\mathrm{RMS}}$ [m/s]
& 0.03828 & 0.04045 & 0.03902 & 0.04974 & \textbf{0.03517} \\
$r_{a,o,\mathrm{RMS}}$ [rad/s]
& 0.12993 & 0.14584 & 0.25285 & 0.15967 & \textbf{0.12621} \\
$I_{\mathrm{push}}$ [$\mathrm{rad}^2/\mathrm{s}$]
& 0.18949 & 0.17946 & 0.53104 & 0.05877 & \textbf{0.02110} \\
$\tau_{\mathrm{RMS}}$ [N\,m]
& 4.1417 & 3.8185 & 3.4817 & 3.5812 & \textbf{3.3582} \\
$\epsilon_{\dot q,\max}$ [rad/s]
& 0.30601 & 0.35773 & 0.58510 & \textbf{0} & \textbf{0} \\
\hline
\end{tabular}
\end{table}

The first two rows of Table~\ref{tab:main_exp_comparison} show that PINV, DLS, and NS-JLA satisfy the instantaneous kinematic equation with almost zero commanded residuals, but this does not give the best realized motion. The proposed method gives the lowest RMS position and orientation errors, maximum position error, and RMS linear and angular realized task-velocity residuals. The only realized tracking metric not minimized by the proposed method is maximum orientation error, where PINV is lower by $5.8\%$. NS-JLA remains close in RMS position error but gives larger orientation error and angular task-velocity residual, so the proposed method provides the more balanced realized response.

The comparison with TP-QP isolates the effect of the objective because both TP-QP and the proposed method satisfy the required-velocity bounds, giving $\epsilon_{\dot q,\max}=0$. Relative to TP-QP, the proposed method reduces RMS position error, RMS orientation error, RMS linear task-velocity residual, and RMS angular task-velocity residual by $29.6\%$, $20.2\%$, $29.3\%$, and $21.0\%$, respectively. Hence, the improvement is attributed not only to admissibility bounds, but also to selecting the admissible required velocity through controller-compatible objective terms that also account for actuator capacities. Relative to PINV, the corresponding reductions are $9.6\%$, $1.5\%$, $8.1\%$, and $2.9\%$, with an additional $22.6\%$ reduction in maximum position error. This comparison is relevant because PINV gives the direct minimum-norm kinematic solution but does not account for the available joint authority near limits.

The limit-pushing and torque metrics further support this interpretation. The proposed method gives the lowest $I_{\mathrm{push}}$, reducing it by $88.9\%$, $88.2\%$, $96.0\%$, and $64.1\%$ relative to PINV, DLS, NS-JLA, and TP-QP, respectively. It also gives the lowest RMS torque norm, reducing it by $18.9\%$, $12.1\%$, $3.5\%$, and $6.2\%$, respectively. These trends are consistent with the controller-compatible objective: among admissible required velocities, the proposed method discourages commands toward nearby limits and penalizes lower-capacity actuator directions while retaining task-relevant motion through redundancy. Since $\boldsymbol{W}_{\tau}$ depends only on known actuator limits and not on measured torque, the torque-norm result is interpreted as supporting evidence rather than a torque-optimality guarantee.


\section{Conclusion}

This paper presented a controller-compatible redundant IK formulation for torque-controlled redundant robots with joint-limit constraints. The IK layer maps the endpoint required twist to joint required velocity through a QP and can serve as an intermediate reference-generation layer for different downstream torque-level controllers. Here, the VDC controller of a 7-DoF upper-limb exoskeleton was kept unchanged to evaluate IK redundancy resolution under the same torque-level control framework. The formulation combines CBF-style reference-level admissibility bounds, a slack-based task relation to avoid infeasibility, previous-command consistency, and actuator torque-capacity weighting, so the selected required velocities remain task-related and more suitable for realization by the torque-level controller. Experiments showed zero required-velocity admissibility violation, a reduced limit-pushing index, lower RMS torque norm, and improved almost all of the realized task-tracking metrics compared with the tested baselines. These results indicate lower limit-pushing behavior, lower RMS torque norm, bounded admissible required velocities, and improved realized task behavior without redesigning the downstream torque controller. The findings should be interpreted as reference-level joint-limit awareness and controller-compatible redundancy resolution, not plant-level safety proof. Future work will investigate coupled IK--torque-control constraint formulations, extension to different torque-level control architectures, and formal closed-loop constraint analysis.

\section*{Acknowledgment}
We acknowledge the financial support of the Finnish Ministry of Education and Culture through the Intelligent Work Machines Doctoral Education Pilot Program (IWM VN/3137/2024-OKM-4).

\bibliography{ref}

\end{document}